%% file: text_detection.tex
\documentclass{ieeeaccess}

\usepackage{amsmath,amssymb,amsfonts}
\usepackage{stfloats}
\usepackage{cite}
\usepackage{graphicx}
\usepackage{psfrag}
\usepackage{subfigure}
\usepackage{amsmath}
\usepackage{array}
\usepackage{algpseudocode}
\usepackage{setspace}
\usepackage{blindtext}
\usepackage{url}
\usepackage{epstopdf}
\usepackage{verbatim}
\usepackage{color}
\usepackage{amsmath}
\usepackage{bm}
\usepackage{algorithm, algorithmicx}
\usepackage{multirow}
\usepackage{booktabs}
\usepackage{makecell}
\usepackage{ntheorem}
\usepackage{textcomp}
\usepackage{caption}
\usepackage{amsmath}
\usepackage{float}
\def\BibTeX{{\rm B\kern-.05em{\sc i\kern-.025em b}\kern-.08em
    T\kern-.1667em\lower.7ex\hbox{E}\kern-.125emX}}
\begin{document}
\history{Date of publication xxxx 00, 0000, date of current version xxxx 00, 0000.}
\doi{10.1109/ACCESS.2017.DOI}
\title{Arbitrary-Shaped Text Detection with Adaptive Text Region Representation}
\author{\uppercase{Xiufeng Jiang}, 
\uppercase{Shugong Xu,
\IEEEmembership{Fellow, IEEE}, Shunqing Zhang,\IEEEmembership{Senior Member, IEEE},and Shan Cao, \IEEEmembership{Member, IEEE}}}
\address[]{Shanghai Institute for Advanced Communication and Data Science, Shanghai University, Shanghai 200444, China (e-mail: \{xiufengjiang, shugong, shunqing, cshan\}@shu.edu.cn)}
\tfootnote{}

\markboth
{X. Jiang \headeretal: Arbitrary-Shaped Text Detection with Adaptive Text Region Representation}
{X. Jiang \headeretal: Arbitrary-Shaped Text Detection with Adaptive Text Region Representation}

\corresp{Corresponding author: Shugong Xu (shugong@shu.edu.cn)}

\begin{abstract}
Text detection/localization, as an important task in computer vision, has witnessed substantial advancements in methodology and performance with convolutional neural networks. However, the vast majority of popular methods use rectangles or quadrangles to describe text regions. These representations have inherent drawbacks, especially relating to dense adjacent text and loose regional text boundaries, which usually cause difficulty detecting arbitrarily shaped text. In this paper, we propose a novel text region representation method, with a robust pipeline, which can precisely detect dense adjacent text instances with arbitrary shapes. We consider a text instance to be composed of an adaptive central text region mask and a corresponding expanding ratio between the central text region and the full text region. More specifically, our pipeline generates adaptive central text regions and corresponding expanding ratios with a proposed training strategy, followed by a new proposed post-processing algorithm which expands central text regions to the complete text instance with the corresponding expanding ratios. We demonstrated that our new text region representation is effective, and that the pipeline can precisely detect closely adjacent text instances of arbitrary shapes. Experimental results on common datasets demonstrate superior performance of our work.
\end{abstract}


\titlepgskip=-15pt

\maketitle

\section{Introduction}
\label{sec:introduction}
As a fundamental task in computer vision, accurate text detection is applicable to many fields in the real world including automatic identity recognition, financial document analysis and recognition, and environmental understanding. In the era of deep learning, the community has witnessed substantial advancements in methodology and performance of text detection. This task, however, is facing many challenges because of various image attributes, such as complex backgrounds, lighting conditions and arbitrary shapes. Our proposed method focuses on the arbitrary-shaped text detection with the same as previous state-of-the-art approaches.

\begin{figure}
\centering
\includegraphics[scale=0.45]{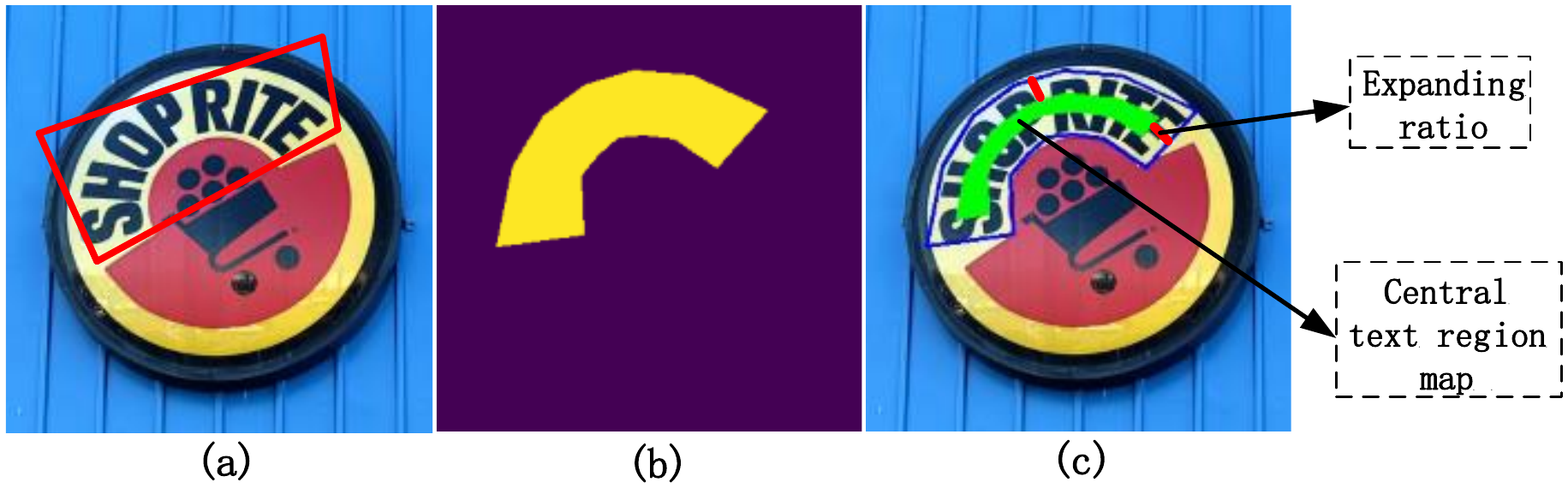}
\caption{Text region representation methods. (a) text region represented by polygon; (b) text region represented by segmentation map; (c) text region represented by central text region map and expanding ratio.}
\label{fig1}
\end{figure} 

Existing scene text detection methods use two forms to represent text regions: quadrangular representation and segmentation mask representation. To the best of our knowledge, most existing regression-based methods are specifically designed to use quadrangular representation to describe text regions, as is the case with general object detection approaches. However, the representation fails to deal with text instances of arbitrary shapes, e.g., the curved texts as shown in Figure \ref{fig1}(a), which contains extraneous information in addition to the text region. Segmentation mask representation as shown in Figure \ref{fig1}(b), on the other hand, are appropriate for arbitrarily shaped text. However, tiny intervals between text regions are very common in natural scenes and it is challenging to separate dense adjacent text instances using this representation. 

To address the above problems, we propose a robust pipeline with novel text representation. We consider a text instance to be composed of a central text region, and an expanding ratio between the central text region and the full text region, of which one example is shown in Figure \ref{fig1}(c). Our proposed method has the following three benefits. First, the central text region mask has a similar shape to the original text instance, which allows the proposed method to represent texts of arbitrary shapes and locate each text instance precisely. Second, the central text region map only captures the central area of each text, which allows separation of spatially close text instances. Third, we propose a post-processing algorithm called ``polygon expansion algorithm", with which the identified full text instances can be successfully expanded from the central text region maps. Moreover, to further improve the representation robustness against text instances of various scales and aspect ratios, we introduce a training strategy that can adaptively adjust the central text region map and the expanding ratio based on multiple expanding ratios of the same text region.

Based on the experiments, the proposed method can achieve 81.3\%, 82.5\%, 84.8\% and 83.6\% F-measure, the state-of-the-art or competitive performance, on DAST1500, CTW1500, TotalText and MSRA-TD500 evaluation datasets, where the proposed novel text representation provides 1.9\%, 0.3\%, 0.1\% and 0.7\% improvement, respectively.The contributions of this paper can be summarized as follows:
\begin{itemize}
\item A novel text region representation method, which can precisely describe dense adjacent text instances with arbitrary shapes.
\item A training strategy to obtain the adaptive text region representations for text instances of various scales and aspect ratios.
\item A polygon expansion algorithm which is able to accurately expand the central text region map to the full text map.
\item Superior performance on representative scene text datasets.
\end{itemize}

\section{Related Work}
As an important research area in computer vision, scene text detection has been inevitably influenced by the wave generated by the deep learning revolution. Most previous deep learning methods can be roughly divided into three categories according to the text region representation: quadrangular representation, segmentation mask representation and hybird methods which predict the segmentation mask of scene texts and regress the bounding boxes of the text instances at the same time.

\textbf{Quadrangular representation} mainly draws inspiration from general object detection frameworks, such as Faster R-CNN\cite{ren2015faster} and SSD\cite{liu2016ssd}. These methods use quadrangles to represent text regions, which can further be divided into one-stage and two-stage approaches, similar to object detection methods. Following Faster R-CNN, CTPN\cite{tian2016detecting} detected long horizontal text by connecting rectangular anchors of consistent width and different heights. To locate multi-oriented text regions, R2CNN\cite{jiang2017r2cnn} refined the axis-aligned boxes and predicted the inclined minimum area boxes with pooled features of different pool sizes. RRPN\cite{ma2018arbitrary} added rotation to both anchors and RoIPooling in Faster R-CNN pipeline. For the one-stage text detector based on SSD, Textboxes\cite{liao2017textboxes} modified anchors and kernels of SSD to detect large-aspect-ratio scene texts. SegLink\cite{shi2017detecting} proposed to predict text segments and then group these segments into text boxes by predicting segments links. To represent curved text regions, TextSnake\cite{long2018textsnake} proposed to regress a sequence of disks with different radii. But it still requires complicated post-processing, and radius regression may result in drop of precision. And as mentioned in the paper\cite{wang2019arbitrary}, Wang proposed the use of polygons of an adaptive number of points to represent text regions. However, the representation is still a polygon and can not describe curved text regions with smooth boundaries.

Most quadrangular representation methods need to carefully design aspect ratios of anchor boxes which depend heavily on experience and may have multiple stages. Moreover, for curved text instances, which are common in application, quadrangular representation is unsatisfactory.

\begin{figure}
\centering
\includegraphics[scale=0.3]{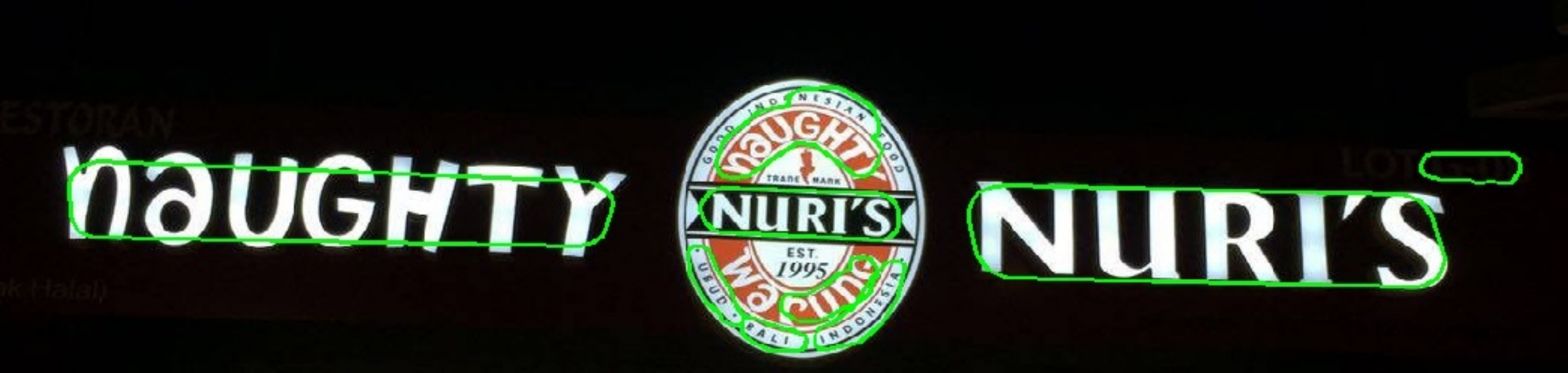}
\caption{The example result of DB\cite{liao2019real}. Detection boundary is inaccurate especially for the text "naUGHTY".}
\label{fig2}
\end{figure}

\textbf{Segmentation mask representation} is mainly based on the classification of each pixel in the image and then clustering to different text instances. To the best of our knowledge, semantic segmentation mask representation is quite suitable for text regions of arbitrary shapes. Zhang et al.\cite{zhang2016multi} detected multi-oriented text by semantic segmentation and MSER-based algorithms. Pixel-Link\cite{deng2018pixellink} performs text/non-text and link prediction on an input image, then adds some post-processing to get text boxes and filter noise. To separate different text instances, PSENet\cite{wang2019shape} outputs various text kernels and uses a progressive scale expansion algorithm to obtain final text boxes. However, it has many output kernels which may have negative effects on location results. \cite{tian2019learning} adopts a mirror symmetry of FPN\cite{lin2017feature} to produce embedding features and text foreground masks, and uses cluster processing to detect texts. DB\cite{liao2019real} proposes a Differentiable Binarization module to predict the shrunk regions, and the shrunk regions are dilated with an constant expanding ratio. However, as shown in Figure \ref{fig2}, using an constant expanding ratio is inaccurate according to the Vatti clipping algorithm\cite{vati1992generic}.

\textbf{Hybrid approaches} are based on quadrangular representation methods and segmentation  mask  representation methods, which predict the segmentation mask of scene texts and regress the bounding boxes of the text instances at the same time. EAST\cite{zhou2017east} and DeepReg\cite{he2017deep} adopt FCNs to predict shrinkable text score maps and perform per-pixel regression, followed by a post-processing NMS. Mask TextSpotter\cite{liao2019mask} detected arbitrary-shape text instances in an instance segmentation manner based on Mask R-CNN.

While quadrangular representation is often not flexible enough, segmentation mask representation fails to separate dense adjacent text instances. In our proposed method, the novel text region representation can flexibly detect dense adjacent text of arbitrary shape, and our proposed Polygon Expansion Algorithm needs only one clean and efficient step.

\section{Proposed method}
This section is organized as follows: we first introduce the overall structure of the proposed method and the novel text region representation, then the training strategy. Next, we illustrate the details of the polygon expansion algorithm. Then, we introduce how to generate the central text region mask and expanding ratios label. Lastly, we explain the loss function design.

\begin{figure*}[ht]
\centering
\includegraphics[width=1\textwidth]{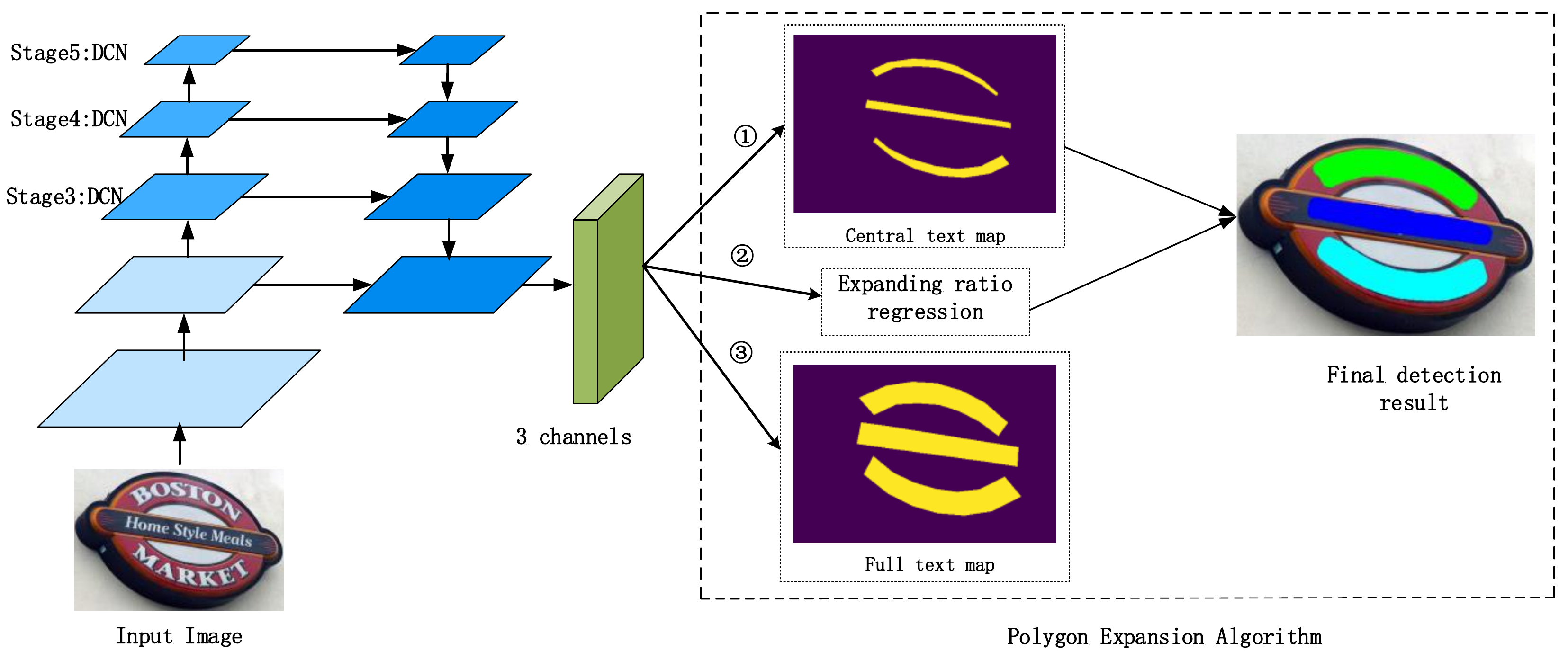}
\caption{Overall architecture of our network designed for arbitrarily shaped text detection. The left part of pipeline is implemented from FPN with deformable convolutional networks. The right part is the polygon expansion algorithm.}
\label{fig3}
\end{figure*}

\subsection{Overall Pipeline}
The overall structure of our proposed network is illustrated in Figure \ref{fig3}. First, ResNet50\cite{he2016deep} is used in our network as the backbone. As mentioned before, text instances in the natural image usually consist of arbitrary shapes. The superior performance of Deformable Convolutional Networks\cite{zhu2019deformable} arises from its ability to adapt to the geometric variation of objects. Therefore, we add deformable convolutional networks(DCN) to the ResNet50 backbone to extract features. Similar to \cite{zhu2019deformable}, we apply deformable convolutional layers at stage3-stage5 of ResNet50. Specifically, we use deformable conv$_{3\times3}$ instead of general conv$_{3\times3}$ in the bottleneck. 

Second, we use the similar feature merging strategy of the U-net\cite{ronneberger2015u} to combine low-level texture features and high-level semantic features. The high-level semantic features make use of upsampling to concatenate with low-level texture features.  Therefore, the fusion feature has various receptive fields which are adaptive to texts of arbitrary shapes. Next, the fusion feature map is projected into three branches to produce complete text segmentation results, along with central text region maps and expanding ratios. After obtaining these outputs, we use the polygon expansion algorithm to expand the central text regions to their complete text regions using the corresponding expanding ratios. 

\subsection{Text region representation}
As mentioned in the introduction section, the text targets are represented in two forms: rectangles or quadrangles, and segmentation mask. The first representation falls short when handling curved texts. The other encounters difficulty separating dense adjacent text instances. Therefore, we propose a new representation for text regions. As shown in Figure \ref{fig1}(c), we apply the central text region map, which has a similar shape to the original text region, and the expanding ratio to represent the text region. On the one hand, the representation combines the advantages of segmentation mask representation, which can precisely describe text regions of arbitrary shapes. On the other hand, the representation solves the problem of separating dense adjacent text regions owing to large geometrical margins among central text regions.

\subsection{Training strategy} 
\begin{figure}[t]
\centering
\includegraphics[scale=0.3]{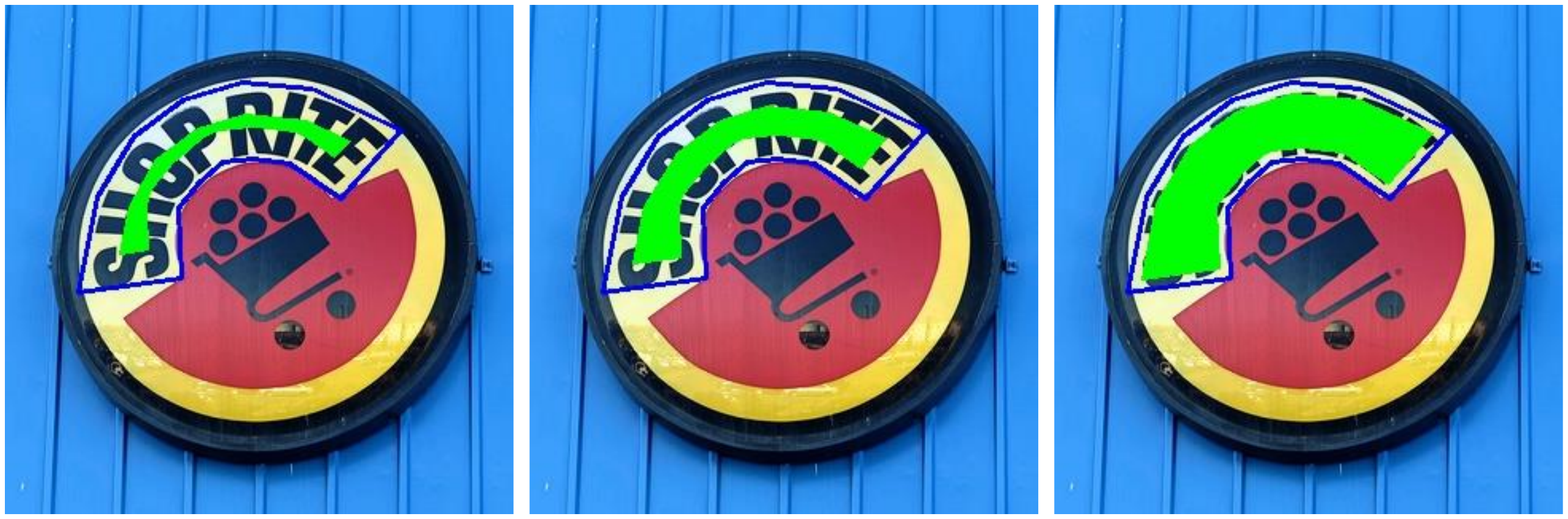}
\caption{Different expanding ratios and corresponding central text region representations for the same text instance.}
\label{fig4}
\end{figure}

For our proposed text region representation method, the representation of each text region is not unique. Specifically, we can represent the same text instance with different expanding ratios and corresponding central text regions as shown in Figure \ref{fig4}. Moreover, multiple text instances can be included in the same image, and their expanding ratios can be different. Therefore we need to find an adaptive representation for each text region. In every training iteration, we propose that our network is trained with different text instances labeled by different expanding ratios and corresponding central text regions of the same image. And another strategy is to use different expanding ratios of the same text instance(e.g. Figure \ref{fig4}) in different training iterations to train the network. With multiple iterations of the network, the network can learn the adaptive representations against text instances of various scales and aspect ratios.

\subsection{Polygon Expansion Algorithm}
As aforementioned, we adopt the central text region map and the expanding ratio to represent the text region. However, our ultimate goal is to detect the entire text region. Therefore we propose a polygon expansion algorithm to obtain the complete text region detection results.

\begin{figure*}[ht]
\centering
\includegraphics[width=1\textwidth]{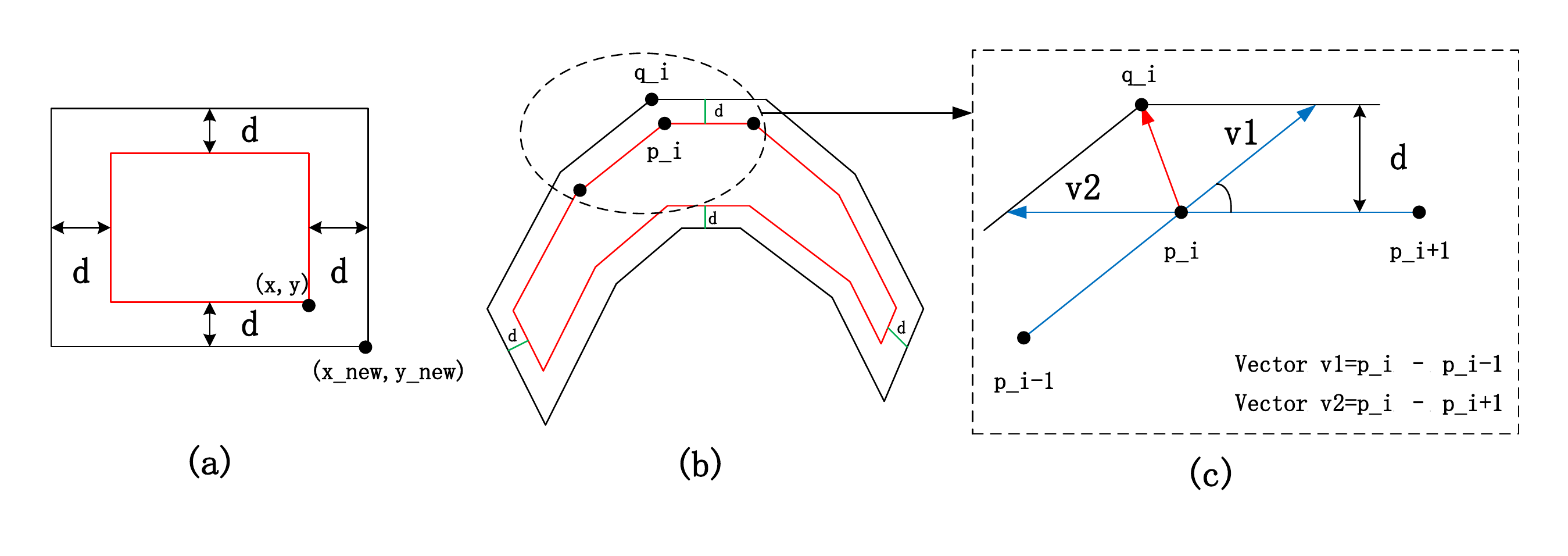}
\caption{The explanation of the polygon expansion algorithm. (a) is the expansion of the rectangle boxes with expanding ratio $d$. (b) refers to the polygon expansion of curved shape with expanding ratio $d$. (c) is the procedure of polygon expansion algorithm.}
\label{fig5}
\end{figure*}

The polygon expansion algorithm is basically an isometric scaling of polygons. Corresponding boundaries of the original and expanded polygon are equidistant. For rectangles such as the one in Figure \ref{fig5}(a), it is simple to derive the new coordinate with the expanding ratio $d$. We can directly obtain the results as follows:
$$(x_{new},y_{new}) = (x,y) \pm d \eqno{(1)}$$

\begin{algorithm}
	\caption{Polygon Expansion Algorithm}
	\label{alg:1}
	\begin{algorithmic}[1]
		\Require Central Text Region Boundary Points-List: $P$; 
		
		\quad Expanding ratio: ${D}$
		\Ensure Full Text Region Points-List: $Q$
		\Function{Expansion}{$P$, $D$}  
            \State $Q \gets \emptyset$ 
            \State $Clockwise$ $P$  
            \For{each $p_{i} \in P$}  
                \State $\Vec{v_{1}} \gets (p_{i} - p_{i-1})$;$\Vec{v_{2}} \gets (p_{i} - p_{i+1})$ 
                \State $\sin<\Vec{v_{1}},\Vec{v_{2}}>$ $\gets (|\Vec{v_{1}} \times \Vec{v_{2}}|\div|\Vec{v_{1}}|\div|\Vec{v_{2}}|)$
                \State $Norm(\Vec{v_{1}}) \gets \frac{\Vec{v_{1}}}{|\Vec{v_{1}}|}$
                \State $Norm(\Vec{v_{2}}) \gets \frac{\Vec{v_{2}}}{|\Vec{v_{2}}|}$
                \State $Norm(\overrightarrow{p_{i}q_{i}})\gets(Norm(\Vec{v_{1}})+Norm(\Vec{v_{2}}))$ 
                \State $\overrightarrow{p_{i}q_{i}}\gets(\frac{D}{\sin<\Vec{v_{1}},\Vec{v_{2}}>}\cdot$$Norm(\overrightarrow{p_{i}q_{i}}))$
                \State $q_{i}\gets(p_{i}+\overrightarrow{p_{i}q_{i}})$
                \State $Q.append(q_{i})$
            \EndFor  
            \State \Return{$Q$}  
        \EndFunction  
	\end{algorithmic}
\end{algorithm}

However, text instances in natural images always have arbitrary shapes. Conveniently, methods like $findContours$ in OpenCV can be applied to obtain the boundaries of central text regions maps. More specifically, by using $findContours$ method, we can obtain the boundary points to represent the central text region map. Moreover, for curved texts, more points are adopted for text region representation as shown in Figure \ref{fig5}(b). The procedure of expanding the central text region to the complete text instance with the corresponding expanding ratio is illustrated in Figure \ref{fig5}(c). It shows how to get one of the full text region points $q_{i}$ with central text region point $p_{i}$ and expanding ratio $d$. First, we need to sort all the central text region boundary points in a clockwise direction to get $P$. Second, we calculate the vectors of point $p_{i}$ and two adjacent points, as $\vec v_{1},\vec v_{2}$. Then we can derive the sine of the angle between $\vec v_{1}$ and $\vec v_{2}$ through their cross product, as formulated in Eqn. (2). Third, we calculate the unit vector of $\vec v_{1}$ and $\vec v_{2}$, to obtain the unit vector of $\overrightarrow{p_{i}q_{i}}$, named as Norm($\overrightarrow{p_{i}q_{i}}$). Thus, Norm($\overrightarrow{p_{i}q_{i}}$) can be formulated as Eqn. (3).
$$\sin<\Vec{v_{1}},\Vec{v_{2}}> = \frac{|\Vec{v_{1}} \times \Vec{v_{2}}|}{|\Vec{v_{1}}|\cdot|\Vec{v_{2}}|} \eqno{(2)}$$
$$Norm(\overrightarrow{p_{i}q_{i}}) = (Norm(\Vec{v_{1}})+Norm(\Vec{v_{2}}))\eqno{(3)}$$ 
Finally, according to its geometric relationship, $\overrightarrow{p_{i}q_{i}}$ can be obtained using the predicted expanding ratio $d$, as shown in Eqn. (4). Therefore, we can get one of the full text region points $q_{i}$ according to $\overrightarrow{p_{i}q_{i}}$. Similarly, we can derive all the final text region points as a result.
$$\overrightarrow{p_{i}q_{i}}=(\frac{d}{\sin<\Vec{v_{1}},\Vec{v_{2}}>})\cdot Norm(\overrightarrow{p_{i}q_{i}})\eqno{(4)}$$

The details of polygon expansion algorithm are summarized in Algorithm 1. In the pseudocode, $P$ is a collection of the central text region boundary points, $D$ is the corresponding expanding ratio and $Q$ represents the detected full text region points.

\subsection{Label Generation}
As illustrated in Figure \ref{fig3}, our network produces central text and full text map, as well as the expanding ratio. Therefore, it requires the corresponding ground truths with segmentation masks and expanding ratios. In our practice, we can conduct these ground truth labels simply and effectively by shrinking the original text labels. As shown in Figure \ref{fig4}, the polygon with blue boundaries is the original text label, which denotes the full text mask label. To obtain the central text mask labels, we use the Vatti clipping algorithm\cite{vati1992generic} to shrink the original text labels. The three green masks in Figure \ref{fig4} are the central text labels with different expanding ratios from origin text label. Subsequently, each central text region and origin text region is transferred into a $0/1$ binary mask for segmentation ground truth labelling. For the expanding ratio lable, similar to the rotation angle labels in EAST\cite{zhou2017east}, we cover the central text region with the expanding ratio value.

\subsection{Loss Function}
For training our network, the loss function can be formulated as:
$$L = \lambda_{1}L_{c} + \lambda_{2}L_{s} + \lambda_{3}L_{d}\eqno{(5)}$$
$$\lambda_{1} + \lambda_{2} + \lambda_{3} = 1\eqno{(6)}$$
where $L_{c}$ represents the loss for the complete text region maps, $L_{s}$ represents the loss for the central ones, and $L_{d}$ represents the expanding ratios regression loss. $\lambda_{i}(i=1,2,3)$ is a hyper-parameter which balances the importance of the different losses.

As we all know, most text instances in natural images only occupy an extremely small region, which makes the predictions of network biased towards the non-text regions, if binary cross entropy\cite{de2005tutorial} is used. To address the imbalance of positive and negative samples, we adopt dice coefficient\cite{milletari2016v} in our experiment. It directly uses the segmentation evaluation protocols as the loss to supervise the network and also ignores a large number of background pixels when calculating the Intersect-over-Union(IOU). The dice coefficient $D(R, G)$ is formulated as follows:
$$D(R, G) = \frac{2\sum_{x,y}(R_{x,y}\times G_{x,y})}{\sum_{x,y}R_{x,y}^2 + \sum_{x,y}G_{x,y}^2}\eqno{(7)}$$
where $R$ and $G$ represent the prediction result and ground truth respectively.

Furthermore, as stated in the paper\cite{shrivastava2016training}, detection tasks usually contain an overwhelming number of easy examples and a small number of hard examples. Automatic selection of these hard examples can make training more effective and efficient. Therefore, we adopt Online Hard Example Mining(OHEM)\cite{shrivastava2016training} to $L_{c}$ during training to get better performance.

$L_{c}$ pays attention to segment the text and non-text regions. We obtain the training mask $M$ using OHEM. Thus $L_{c}$ can be formulated as:
$$L_{c} = 1-D(R_{c}\cdot M, G_{c}\cdot M)\eqno{(8)}$$

$L_{s}$ is the loss for central text regions. Like the paper PSENet\cite{wang2019shape}, we ignore the pixels of the non-text regions in the segmentation result $R_{c}$ to avoid redundancy. Therefore, $L_{s}$ is formulated as:
$$L_{s} = 1-D(R_{s}\cdot W, G_{s}\cdot W)\eqno{(9)}$$
$$W=
\begin{cases}
1& \text{$if$ $R_{c} \ge0.5$}\\
0& \text{$otherwise$}
\end{cases}\eqno{(10)}$$
where $W$ is a mask which ignores the pixels of the non-text region in $R_{c}$.

$L_{d}$ is the regression loss of the expanding ratio, which is the key part to obtain the final detection from the central text region. We employ pixel-wise smooth $L_{1}$ loss\cite{girshick2015fast} to optimize the loss. The loss is formulated as:
$$L_{d} = \sum_{x,y}smooth_{L_{1}}(P_{x,y}-Q_{x,y})\eqno{(11)}$$
in which
$$smooth_{L_{1}}(x)=
\begin{cases}
0.5x^2& \text{$if$ $|x|<1$}\\
|x|-0.5& \text{$otherwise$}
\end{cases}\eqno{(12)}$$
where $P$ and $Q$ represent the regression result and ground truth, respectively.

\section{Experiment}
In this section, we first briefly introduce datasets and explain the implementation details. Then we discuss our proposed method and conduct ablation studies. Finally, five standard benchmarks are used in this paper for performance evaluation: TotalText, CTW1500, MSRA-TD500 ICDAR 2015 and DAST1500, achieving on par or better results than state-of-the-art methods.

\subsection{Benchmark Datasets}
\textbf{ICDAR 2017 MLT}\cite{nayef2017icdar2017} is a large scale multi-lingual text dataset, which consists of 9 languages. It contains three parts: 7200 training images, 1800 validation images, and 9000 test images. The text regions are annotated by 4 vertices of the word quadrangle. We use both training set and validation set to pretrain our model.

\textbf{ICDAR 2015}\cite{karatzas2015icdar} is a dataset which focuses on natural scene text images. Text regions are multi-oriented. The benchmark consists of 1500 images, 1000 of which are used for training, and the remaining for testing. Image annotations are labelled as word-level quadrangles. In our training, we simply ignore instances which suffered from motion blur and other problems, marking them as `DO NOT CARE'.

\textbf{CTW1500}\cite{yuliang2017detecting} is a commonly used dataset for challenging long curve text detection. It contains a total of 1500 images: 1000 for training and 500 for testing. In contrast to multi-oriented text instances labels(e.g. ICDAR 2015, ICDAR 2017 MLT), the text regions in CTW1500 are annotated using polygons with 14 vertices.

\textbf{MSRA-TD500}\cite{yao2012detecting} consists of 300 training images and 200 test images collected from natural scenes. It is a multi-language dataset, which includes English and Chinese. The scene texts have arbitrary orientations, which are labeled with inclined boxes made up by 4 points at sentence level. There are some long straight text lines.

\textbf{TotalText}\cite{ch2017total} is a newly-released dataset which focuses on the text of various shapes. This dataset includes horizontal, multi-oriented and curve text instances. There are 1255 training images and 300 testing images.  The scene texts in these images are labeled at word level with adaptive number of corner points.

\textbf{DAST1500}\cite{tang2019seglink++} is a dense and arbitrary-shaped text detection dataset, which contains 1538 images and 45,963 line-level detection annotations (including 7441 curved bounding boxes). The images are manually collected from the Internet and of size around 800 × 800. This dataset is multi-lingual, including mostly Chinese, and few English and Japanese texts. The images are divided as follows: 1038 images for training and 500 images for testing.

Following the ICDAR evaluation protocol, we evaluate our proposed method in terms of Recall, Precision and F-measure. Recall is the fraction of correct detection regions among all the text regions in the dataset while Precision represents the fraction of correct detection regions among all the detection regions. F-measure considers both the recall and precision to compute the score. A correct detection region is that the overlap between prediction and ground truth is larger than a given threshold. The computation of the three evaluation terms is usually different for different datasets. For ICDAR datasets, we can evaluate by ICDAR robust reading competition platform, and other datasets can be evaluated with the given evaluation methods corresponding to them.

\subsection{Implementation Details}
\textbf{Training} The backbone of our network is ResNet50\cite{he2016deep} with deformable convolutional networks(dcn)\cite{zhu2019deformable}, where we replace $conv_{3\times3}$ with $dcn_{3\times3}$ in stage3-stage5 of ResNet50. The network layers are initialized with a pretrained ResNet50 model for ImageNet\cite{deng2009imagenet} classification, while other new layers in our network are randomly initialized via He's method\cite{he2015delving}. For ICDAR 2017 MLT, we use 7200 training set and 1800 validation set to train our network. We use stochastic gradient descent(SGD) to optimize our network. While we train the network with a batchsize of 8 for 600 epochs, the initial learning rate is set to $1\times10^{-3}$ and is divided by 10 at 200 epochs and 400 epochs. We first adopted the warmup strategy in \cite{peng2018megdet}, then we found that without warmup the net can still converge quickly. We use a weight decay of $5\times10^{-4}$ and a Nesterov momentum\cite{sutskever2013importance} of 0.99 without dampening. Our implementation also includes batch normalization\cite{ioffe2015batch} and OHEM\cite{shrivastava2016training}, whose ratio of positive and negative samples is 1 : 3.

There are two training strategies for the rest of the datasets: training from scratch and fine-tuning on IC17-MLT model. When training from scratch, the training sets are the same as above. For fine-tuning on the IC17-MLT model, we train the network with 300 epochs. The initial learning rate is $1\times10^{-4}$ and is divided by 10 at 100 and 200 epochs.

\textbf{Data Augmentation} is used in our training. We first randomly rescale the input image with ratios of [0.5,1.0,2.0,3.0]. Then random rotations, transpositions, and flipping are performed. Finally, $640\times640$ samples are randomly cropped from the transformed images.

\textbf{Post Processing} For quadrangular texts like those of ICDAR 2015, we can use $minAreaRect$ in OpenCV to get the central text boundaries and then our proposed Polygon Expansion Algorithm to obtain the final detection result. While for curved text(e.g., CTW1500), methods like $findContours$ in OpenCV can be applied to obtain the boundaries of central text regions and Polygon Expansion Algorithm to finish text detection task.

\subsection{Ablation Study}
To verify the effectiveness of our proposed method, we do a series of comparative experiments on the test set in the ICDAR 2015.

\textbf{Influence of the full text region map.} As shown in Figure \ref{fig3}, in order to get the final detection result using the polygon expansion algorithm at the inference step, we only need the central text region map and the expanding ratio. Therefore, it is unnecessary to predict the full text region map. However, if we only predict the central text map without a full text map prediction, the detection performance is unsatisfactory as illustrated in Table \ref{t1}. The models are evaluated on ICDAR 2015 dataset. We can find from Table \ref{t1} that the F-measures on the test sets drop a lot when derived without a full text map prediction. The central text region map loses a lot of positive sample information, while text instances in natural images only occupy an extremely small region. Without the full text map, this will cause the imbalance of positive and negative samples to more seriously impact results. 

\begin{table}[t]
\centering
\resizebox{.95\columnwidth}{!}{
\smallskip\begin{tabular}{|l|l|l|l|l|}
\hline
Method & Ext & Recall[\%] & Precision[\%] & F-measure[\%] \\
\hline
Ours(W/O) & - & 72.98 & 77.38 & 75.12 \\
\hline
Ours & - & \textbf{79.68} & \textbf{85.79} & \textbf{82.62} \\
\hline
\end{tabular}
}
\caption{Ablation study on the full text map. ``W/O'' represents the result without the full text map prediction. ``Ext'' indicates external data.}
\label{t1}
\end{table}

\begin{table*}[t]
\centering
\begin{tabular}{|c|c|c|c|c|c|c|c|}
\hline
\multirow{2}*{Loss} & \multirow{2}*{Ext} &\multicolumn{3}{c|}{Task Weights}& F-Seg & C-Seg & Ratio-R \\
\cline{3-5}
 & & F-Seg & C-Seg & Ratio-R & IoU[\%] & IoU[\%] & Mean Error \\
\hline
F-Seg only & - & 1 & 0 & 0 & 86.4 & - & - \\
\hline
C-Seg only & - & 0 & 1 & 0 & - & 84.1 & - \\
\hline
Ratio-R only & - & 0 & 0 & 1 & - & - & 0.08 \\
\hline
Unweighted sum of losses & - & 0.333 & 0.333 & 0.333 & 87.1 & 85.3 & 0.05 \\
\hline
weighted sum of losses & - & 0.25 & 0.5 & 0.25 & 88 & 87.2 & 0.035 \\
\hline
Optimal weights & - & 0.5 & 0.25 & 0.25 & \textbf{90.7} & \textbf{88.9} & \textbf{0.02} \\
\hline
\end{tabular}
\caption{Ablation study on the multiple task. ``F-Seg'' represents the full text region map segmentation task, ``C-Seg'' is the center text map segmentation task and ``Ratio-R'' refers to the expanding ratio regression task. ``Ext'' indicates external data. Results are shown from the ICDAR 2015 test set. We observe an improvement in performance when training with multi-task loss.}
\label{t2}
\end{table*}

\textbf{Influence of the multiple task.} We can find from Figure. \ref{fig3} that our pipeline is a multi-task network, which includes both expanding ratio regression and pixel classification tasks. In one deep multi-task network, which produces multiple predictive outputs, can offer faster speed and better performance than their single-task counterparts. Such network makes multiple inferences in one forward pass, therefore the speed is faster. And on the other hand, the model shares weights across multiple tasks, which can induce more robust regularization and boost performance as a result. We use ICDAR 2015 to do the experiments. Table \ref{t2} shows the experimental results, from which we can find that with multi-task learning, the performance of each single task is more satisfactory with the test set.

\begin{figure}[ht]
\centering
\includegraphics[scale=0.25]{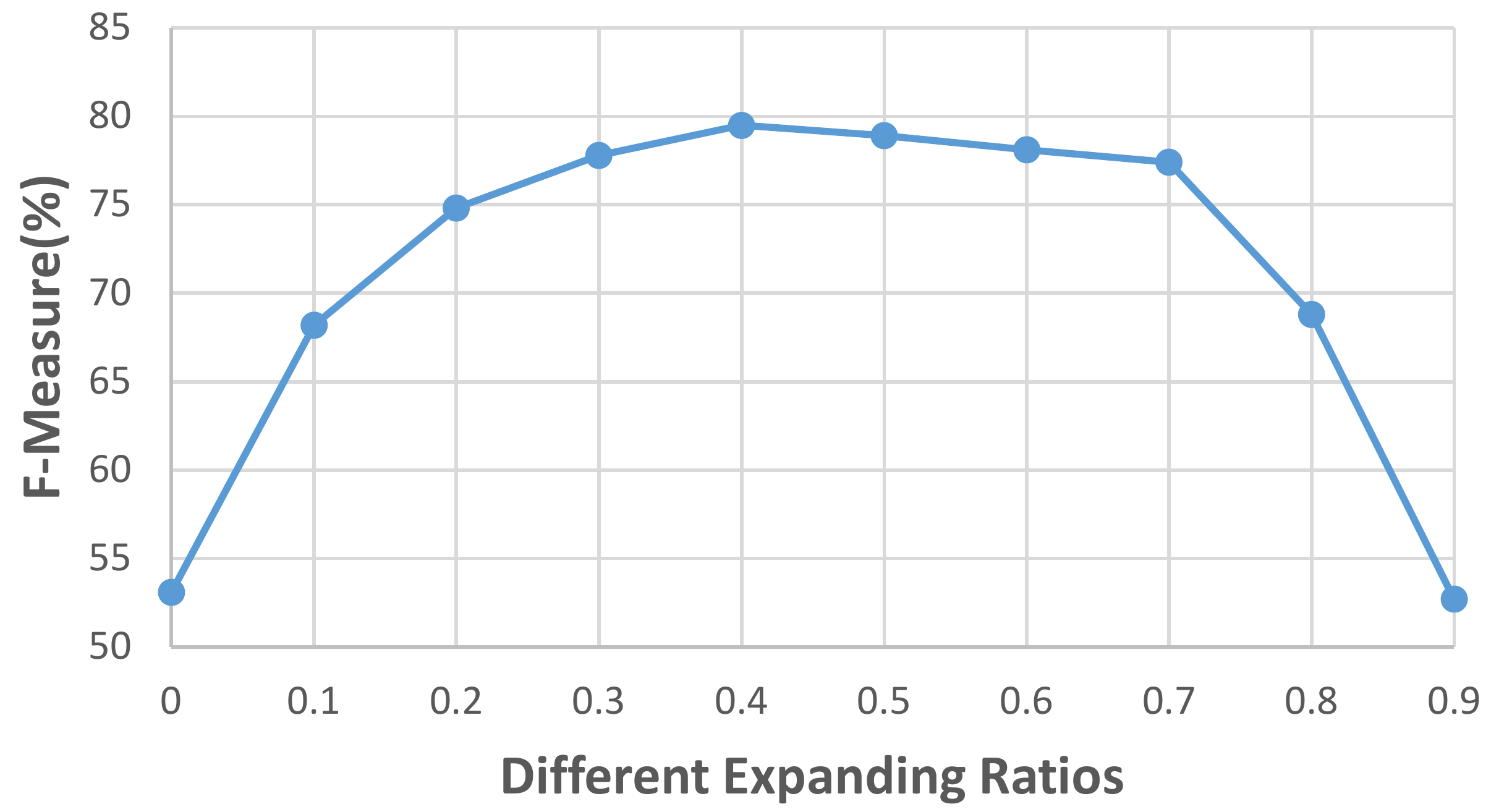}
\caption{Ablation study on different fixed expanding ratios.}
\label{fig6}
\end{figure}

\textbf{Influence of the training strategy.} We investigate the effect of the fixed expanding ratio and the multiple expanding ratios on the performance of our method. As shown in Figure \ref{fig3}, we use the polygon expansion algorithm to expand the central text region map by the expanding ratio. However, if we fix the expanding ratio(e.g. 0.5) to generate central text maps for training, there is no need to regress the expanding ratio. We can directly expand the central text map with the fixed expanding ratio. However, as explained in \cite{tian2019learning}, it may fail in the following case: boxes dilated from central areas with fixed expanding ratio are sometimes smaller or larger than the ground truth boxes. The models are evaluated on ICDAR 2015 dataset. We can find from Figure \ref{fig6} that the F-measures on the test sets drop when the fixed expanding ratio is too large or too small. When the fixed expanding ratio is too small, separating the text instances lying closely to each other is difficult. While the fixed expanding ratio is too large, the predicted central mask is sometimes split into several parts. It is obvious that when the fixed expanding ratio is $0.4$, the performance on the test sets is best and the F-measure is $79.5\%$. Note that when setting the fixed expanding ratio to 0, we only use full text segmentation maps as the final result without the polygon expansion algorithm. But when we use multiple expanding ratios to train the network, the performance of the method with multiple expanding ratios is much higher than with a fixed expanding ratio ($82.62\%$ vs $79.54\%$). Table \ref{t3} shows the experimental results of the multiple expanding ratios method and the fixed expanding ratio(0.4) method. It justifies that the multiple expanding ratios training strategy is more accurate to get complete text regions.

\begin{table*}[t]
\centering
\smallskip\begin{tabular}{|l|l|l|l|l|l|l|l|}

\hline
\multirow{2}*{Method} & \multirow{2}*{Ext} & \multicolumn{3}{c|}{IC15} & \multicolumn{3}{c|}{CTW1500}  \\
\cline{3-8}
 & & Recall[\%] & Precision[\%] & F-measure[\%] & Recall[\%] & Precision[\%] & F-measure[\%] \\
\hline
Ours(Fixed) & - & 76 & 83.41 & 79.54 & 74 & 79.1 & 76.5 \\
\hline
Ours & - & \textbf{79.68} & \textbf{85.79} & \textbf{82.62} & \textbf{75.9} & \textbf{80.6} & \textbf{78.2} \\
\hline
\end{tabular}
\caption{Ablation study on the training strategy. ``Fixed'' refers to a constant expanding ratio of 0.4. ``Ext'' indicates external data.}
\label{t3}
\end{table*}

\begin{figure*}[ht]
\centering

\subfigure[]{
\begin{minipage}[t]{0.95\linewidth}
\centering
\includegraphics[width=1.0\columnwidth]{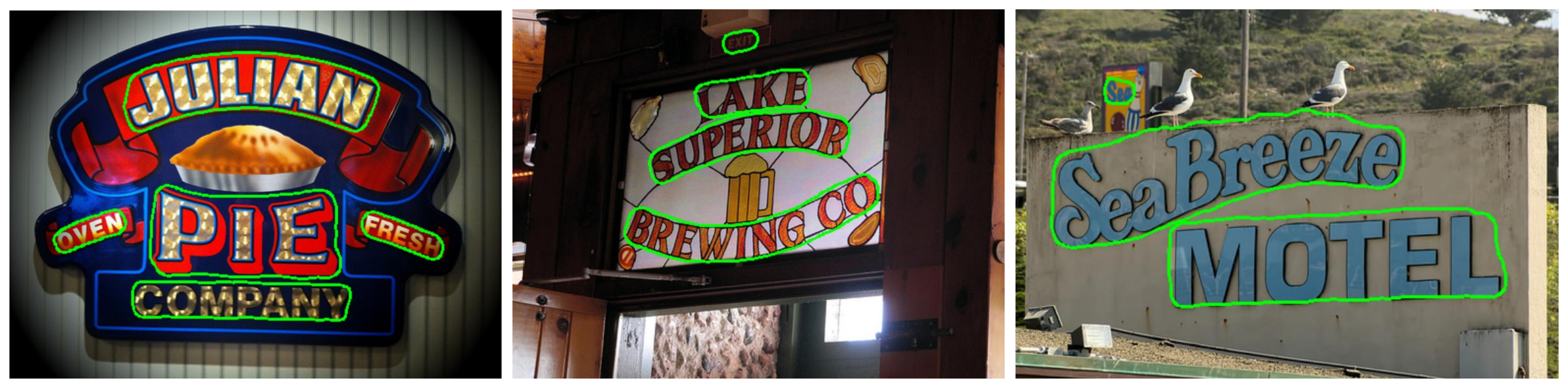}
\label{ff1}
\end{minipage}}

\subfigure[]{
\begin{minipage}[t]{0.95\linewidth}
\centering
\includegraphics[width=1.0\columnwidth]{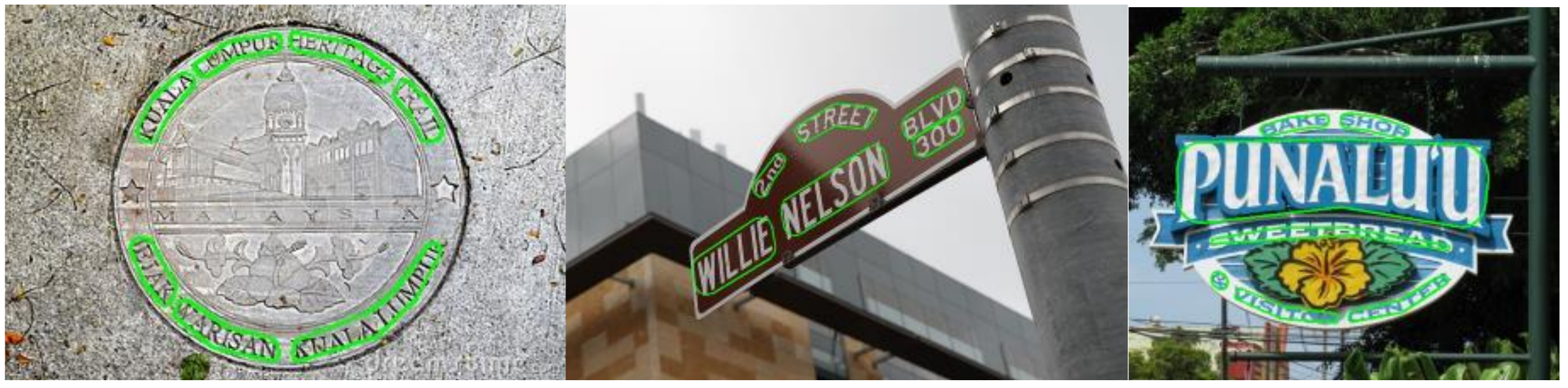}
\label{ff2}
\end{minipage}}

\centering
\caption{Detection results on different datasets. (a) results on CTW1500; (b) results on TotalText.}
\label{fig7}
\end{figure*}

\textbf{Influence of the post-processing.} We verify the effectiveness of our proposed post-processing method on the detection performance and speed. As shown in Table \ref{t4}, our method performs much faster than the previous leading method. PSE\cite{wang2019shape} has many output kernels to merge which may have negative effects on the speed. Notably, our post-processing method outperforms the previous method by 2\%. The comparisons clearly demonstrate that our post-processing method is simple and efficient.

\begin{table}[t]
\centering
\resizebox{1.0\columnwidth}{!}{
\smallskip\begin{tabular}{|l|l|l|l|}
\hline
Method & Ext & F-measure[\%] & Time consumption[ms] \\
\hline
PSE\cite{wang2019shape} & - & 80.6 & 145 \\
\hline
Ours & - & \textbf{82.6} & \textbf{81} \\
\hline
\end{tabular}
}
\caption{Performance and speed with different post-processing on IC15. ``Ext'' indicates external data.}
\label{t4}
\end{table}

\textbf{Influence of the backbone.}
In our proposed method, ResNet50 with deformable convolutional networks(DCN) is the backbone, while ResNet50 is usually used to extract deep features through other state-of-the-art methods. To better analyze the effectiveness of the backbone network, we test the proposed method with different backbone networks(ResNet50, ResNet101, ResNet152 and ResNet50-DCN) on the ICDAR 2015 test set. As shown in Table \ref{t5}, under the same settings, the results show that ResNet50-DCN is better than others. Deformable convolutional networks can clearly improve performance.

\begin{table}[t]
\centering
\resizebox{1.0\columnwidth}{!}{
\smallskip\begin{tabular}{|l|l|l|l|l|}
\hline
Method & Ext & Recall[\%] & Precision[\%] & F-measure[\%] \\
\hline
Ours(ResNet50) & - & 77.99 & 83.89 & 80.83 \\
\hline
Ours(ResNet50-DCN) & - & \textbf{79.68} & \textbf{85.79} & \textbf{82.62} \\
\hline
Ours(ResNet101) & - & 78.87 & 84.64 & 81.65 \\
\hline
Ours(ResNet152) & - & 79.34 & 84.95 & 82.04 \\
\hline
\end{tabular}
}
\caption{Performance with different backbones on IC15. ``Ext'' indicates external data.}
\label{t5}
\end{table}

\subsection{Comparison with State-of-the-art methods}
To show the effectiveness of our proposed approach for datasets of different types, we do a series of tests on several benchmarks. We first evaluate our method on CTW1500 and TotalText, which contain challenging multi-oriented and curved texts. We compare its performance with state-of-the-art ones on CTW1500 and TotalText. Next, to test its ability for oriented text detection, we compare the methods on the widely used benchmark: ICDAR2015. Then, to test the robustness of the proposed method to multi-language and long straight text, we evaluate the method on the MSRA-TD500 dataset. Finally, to show that the proposed method works well for dense adjacent text lines in the natural scene images, we evaluate the method on the DAST1500 dataset.

\begin{figure*}[ht]
\centering

\subfigure[]{
\begin{minipage}[t]{0.95\linewidth}
\centering
\includegraphics[width=1.0\columnwidth]{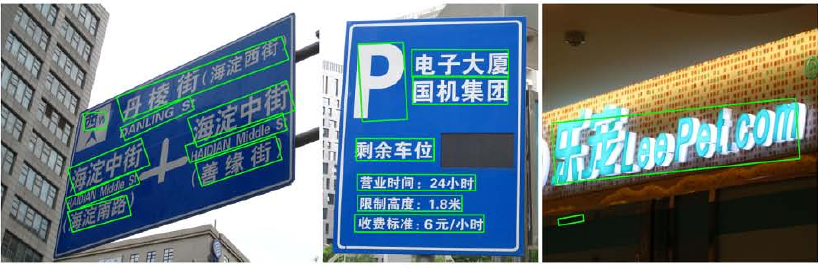}
\label{ff3}
\end{minipage}}

\subfigure[]{
\begin{minipage}[t]{0.95\linewidth}
\centering
\includegraphics[width=1.0\columnwidth]{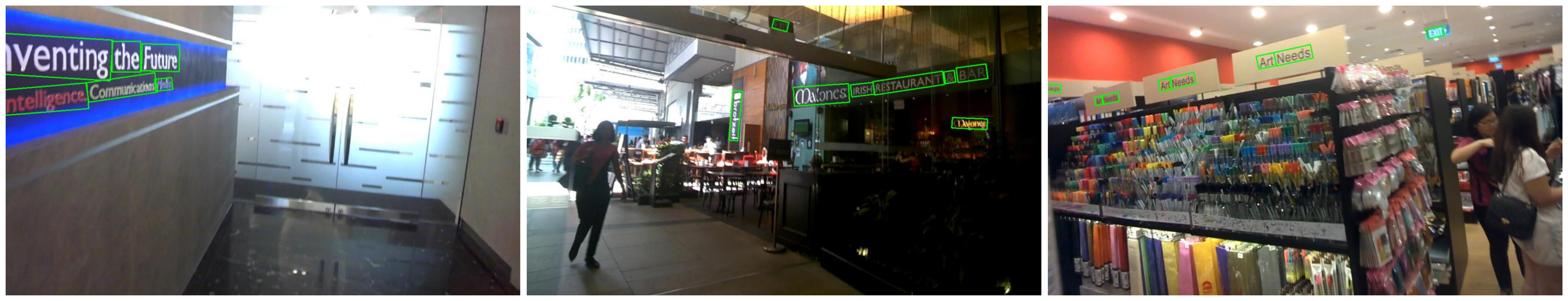}
\label{ff4}
\end{minipage}}

\centering
\caption{Detection results on different datasets. (a) results on MSRA-TD500; (b) results on ICDAR2015.}
\label{fig8}
\end{figure*}

\textbf{Curved text detection}. As shown in Table \ref{t6}, we compare the proposed method with state-of-the-art methods on CTW1500. We have two training strategies in the experiment as explained in the Implementation Details section: training from scratch and fine-tuning on a training model with extra data.  Notably, when other methods and our method train with no extra data, our method outperforms the previous state-of-the-art method by 0.2\%. To our knowledge, this is the best reported result in literature. When other methods and our method train with extra data, our method can achieve competitive performance. While DB adopts differentiable binarization thus it works better on some of the datasets. The performance on the CTW1500 dataset demonstrates the proposed method's superior ability to handle with texts with arbitrary shapes. The comparisons clearly demonstrate that our method can solve the adjacent text instances problem of segmentation-based methods.

Similar conclusions can be obtained on Total-Text. Without external data pre-training, the performance of the method is still very competitive (82.2\%) from Table \ref{t7}. With external data pre-training, the F-measure boosting to 84.8\%, which outperforms the previous state-of-the-art method by 1.2\%. The performance on CTW1500 and Total-Text demonstrates the solid superiority of the proposed method to detect arbitrary-shaped text.

\begin{table}[t]
\centering
\resizebox{.95\columnwidth}{!}{
\smallskip\begin{tabular}{|l|l|l|l|l|}
\hline
Method & Ext & R[\%] & P[\%] & F[\%] \\
\hline
CTPN\cite{tian2016detecting}& - & 53.8 & 60.4 & 56.9 \\
\hline
SegLink\cite{shi2017detecting}& - & 40.0 & 42.3 & 40.8 \\
\hline
EAST\cite{zhou2017east}& - & 49.1 & 78.7 & 60.4 \\
\hline
CTD+TLOC\cite{yuliang2017detecting}& - & 69.8 & 77.4 & 73.4 \\
\hline
PSENet\cite{wang2019shape} & - & 75.6 & \textbf{80.6} & 78.0 \\
\hline
\textbf{Ours} & - & \textbf{75.9} & \textbf{80.6} & \textbf{78.2} \\
\hline
TextSnake\cite{long2018textsnake} & \checkmark & \textbf{85.3} & 67.9 & 75.6 \\
\hline
Wang et al.\cite{wang2019arbitrary} & \checkmark & 80.2 & 80.1 & 80.1 \\
\hline
Tian et al.\cite{tian2019learning} & \checkmark & 77.8 & 82.7 & 80.1 \\
\hline
PSENet\cite{wang2019shape} & \checkmark & 79.7 & 84.8 & 82.2 \\
\hline
DB\cite{wang2019shape} & \checkmark & 80.2 & \textbf{86.9} & \textbf{83.4} \\
\hline
\textbf{Ours} & \checkmark & 80.3 & 84.9 & 82.5 \\
\hline
\end{tabular}
}
\caption{The single-scale results on CTW1500. ``R'', ``P'' and ``F'' represent the recall, precision, and F-measure respectively. ``Ext'' refers to external data.}\smallskip
\label{t6}
\end{table}

\begin{table}[t]
\centering
\resizebox{.95\columnwidth}{!}{
\smallskip\begin{tabular}{|l|l|l|l|l|}
\hline
Method & Ext & R[\%] & P[\%] & F[\%] \\
\hline
SegLink\cite{shi2017detecting}& - & 23.8 & 30.3 & 26.7 \\
\hline
EAST\cite{zhou2017east}& - & 36.2 & 50.0 & 42.0 \\
\hline
DeconvNet\cite{ch2017total}& - & 40.0 & 33.0 & 36.0 \\
\hline
TextSnake\cite{long2018textsnake} & \checkmark & 74.5 & 82.7 & 78.4 \\
\hline
Wang et al.\cite{wang2019arbitrary} & \checkmark & 76.2 & 80.9 & 78.5 \\
\hline
MTS\cite{liao2019mask} & \checkmark & 75.6 & 82.5 & 78.6 \\
\hline
TextField\cite{xu2019textfield} & \checkmark & 79.9 & 81.2 & 80.6 \\
\hline
PSENet\cite{wang2019shape} & \checkmark & 78.0 & 84.0 & 80.9 \\
\hline
SPCNet\cite{xie2019scene} & \checkmark & \textbf{82.8} & 83.0 & 82.9 \\
\hline
LOMO\cite{zhang2019look} & \checkmark & 79.3 & 87.6 & 83.3 \\
\hline
CRAFT\cite{baek2019character} & \checkmark & 79.9 & 87.6 & 83.6 \\
\hline
DB\cite{wang2019shape} & \checkmark & 82.5 & 87.1 & 84.7 \\
\hline
\textbf{Ours} & \checkmark & 80.9 & \textbf{89.0} & \textbf{84.8} \\
\hline
\end{tabular}
}
\caption{The single-scale results on TotalText. ``R'', ``P'' and ``F'' represent the recall, precision, and F-measure respectively.
``Ext'' refers to external data.}\smallskip
\label{t7}
\end{table}

\textbf{Multi-oriented text detection.} Although our method is segmentation-based, which is specifically designed for text detection of curved shapes, our approach is also adaptive to oriented text detection. The detection performance of our method and other state-of-the-art methods on IC15 is given in Table \ref{t8}. Similarly to the CTW1500 training methods and settings, there are two training strategies: training from scratch and fine-tuning on a training model with extra data. When training from scratch, we can find that the performance of our method surpasses the state of the art results by more than 0.1\%. Moreover, for fine-tuning on a training model with extra data, our approach achieves competitive performance with state-of-the-art methods, which means it can also process multi-oriented text well.

\begin{table}[ht]
\centering
\resizebox{0.95\columnwidth}{!}{
\begin{tabular}{|l|l|l|l|l|}
\hline
Method & Ext & R[\%] & P[\%] & F[\%] \\ 
\hline
CTPN\cite{tian2016detecting}& - & 51.6 & 74.2 & 60.9 \\
\hline
RRPN\cite{ma2018arbitrary}& - & 73.0 & 82.0 & 77.0 \\
\hline
EAST\cite{zhou2017east}& - & 73.5 & 83.6 &  78.2 \\
\hline
PSENet\cite{wang2019shape} & - & 79.7 & 81.5 & 80.6 \\
\hline
DeepReg\cite{he2017deep} & - & 80.0 & 82.0 & 81.0 \\
\hline
PixelLink\cite{deng2018pixellink} & - & \textbf{81.7} & 82.9 & 82.3 \\
\hline
R2CNN\cite{jiang2017r2cnn} & - & 79.7 & 85.6 & 82.5 \\
\hline
\textbf{Ours} & - & 79.7 & \textbf{85.8} & \textbf{82.6} \\ 
\hline
SegLink\cite{shi2017detecting}& \checkmark & 76.8 & 73.1 & 75.0 \\
\hline
SSTD\cite{he2017single}& \checkmark & 73.9 & 80.2 & 76.9 \\
\hline
WordSup\cite{hu2017wordsup}& \checkmark & 77.0 & 79.3 & 78.2 \\
\hline
Lyu et al.\cite{lyu2018multi}& \checkmark & 70.7 & \textbf{94.1} & 80.7 \\
\hline
RRD\cite{liao2018rotation}& \checkmark & 79.0 & 85.6 & 82.2 \\
\hline
TextSnake\cite{long2018textsnake} & \checkmark & 80.4 & 84.9 & 82.6 \\
\hline
PSENet\cite{wang2019shape} & \checkmark & 84.5 & 86.9 & 85.7 \\
\hline
MTS\cite{liao2019mask} & \checkmark & 81.0 & 91.6 & 86.0 \\
\hline
Tian et al.\cite{tian2019learning} & \checkmark & 85.0 & 88.3 & 86.6 \\
\hline
SPCNet\cite{xie2019scene} & \checkmark & 85.8 & 88.7 & 87.2 \\
\hline
DB\cite{wang2019shape} & \checkmark & 83.2 & 91.8 & 87.3 \\
\hline
Wang et al.\cite{wang2019arbitrary} & \checkmark & \textbf{86.0} & 89.2 & 87.6 \\
\hline
FOTS\cite{liu2018fots} & \checkmark & 85.17 & 91.0 & \textbf{87.99} \\
\hline
\textbf{Ours} & \checkmark & 83.6 & 86.9 & 85.2 \\
\hline
\end{tabular}
}
\caption{The single-scale results on IC15. ``R'', ``P'' and ``F'' represent the recall, precision, and F-measure respectively. ``Ext'' refers to external data.}
\label{t8}
\end{table}

\begin{table}[ht]
\centering
\resizebox{0.95\columnwidth}{!}{
\begin{tabular}{|l|l|l|l|l|}
\hline
Method & Ext & R[\%] & P[\%] & F[\%] \\ 
\hline
RRPN\cite{ma2018arbitrary}& - & 68.0 & 82.0 & 74.0 \\
\hline
DeepReg\cite{he2017deep} & - & 70.0 & 77.0 & 74.0 \\
\hline
EAST\cite{zhou2017east}& - & 67.4 & \textbf{87.3} &  76.1 \\
\hline
\textbf{Ours} & - & \textbf{77.0} & 80.8 & \textbf{78.9} \\ 
\hline
SegLink\cite{shi2017detecting}& \checkmark & 70.0 & 86.0 & 77.0 \\
\hline
PixelLink\cite{deng2018pixellink} & \checkmark & 73.2 & 83.0 & 77.8 \\
\hline
TextSnake\cite{long2018textsnake} & \checkmark & 73.9 & 83.2 & 78.3 \\
\hline
RRD\cite{liao2018rotation}& \checkmark & 73.0 & 87.0 & 79.0 \\
\hline
Lyu et al.\cite{lyu2018multi}& \checkmark & 76.2 & 87.6 & 81.5 \\
\hline
Tian et al.\cite{tian2019learning} & \checkmark & 81.7 & 84.2 & 82.9 \\
\hline
CRAFT\cite{baek2019character} & \checkmark & 78.2 & 88.2 & 82.9 \\
\hline
DB\cite{wang2019shape} & \checkmark & 79.2 & \textbf{91.5} & \textbf{84.9} \\
\hline
\textbf{Ours} & \checkmark & \textbf{83.1} & 84.2 & 83.6 \\
\hline
\end{tabular}
}
\caption{The single-scale results on MSRA-TD500. ``R'', ``P'' and ``F'' represent the recall, precision, and F-measure respectively. ``Ext'' refers to external data.}
\label{t9}
\end{table}

\begin{table}[ht]
\centering
\resizebox{0.95\columnwidth}{!}{
\begin{tabular}{|l|l|l|l|l|}
\hline
Method & R[\%] & P[\%] & F[\%] \\ 
\hline
TextBoxes++\cite{liao2018textboxes++} & 40.9 & 67.3 & 50.9 \\
\hline
RRD\cite{liao2018rotation} & 43.8 & 67.2 & 53.0 \\
\hline
EAST\cite{zhou2017east} & 55.7 & 70.0 &  62.0 \\
\hline
SegLink\cite{shi2017detecting} & 64.7 & 66.0 & 65.3 \\
\hline
CTD+TLOC\cite{yuliang2017detecting} & 60.8 & 73.8 & 66.6 \\
\hline
PixelLink\cite{deng2018pixellink} & 75.0 & 74.5 & 74.7 \\
\hline
SegLink++\cite{tang2019seglink++} & 79.2 & 79.6 & 79.4 \\
\hline
\textbf{Ours} & \textbf{80.5} & \textbf{82.1} & \textbf{81.3} \\
\hline
\end{tabular}
}
\caption{The single-scale results on DAST1500. ``R'', ``P'' and ``F'' represent the recall, precision, and F-measure respectively.}
\label{t10}
\end{table}

\textbf{Multi-language and long straight text detection.} Our method is adaptive to multi-language and long straight text detection. As shown in Table \ref{t9}, the F-measure of the proposed method is 78.9\% and 83.6\% when the external data is not used and used respectively. For the F-measure,  the method surpasses the previous state-of-the-art method by 2.8\% when training without extra data. When training with extra data, the F-measure is comparable with previous best performance. Thus, our method can indeed be deployed in complex natural scenarios.

\begin{figure*}[ht]
\centering
\includegraphics[width=1\textwidth]{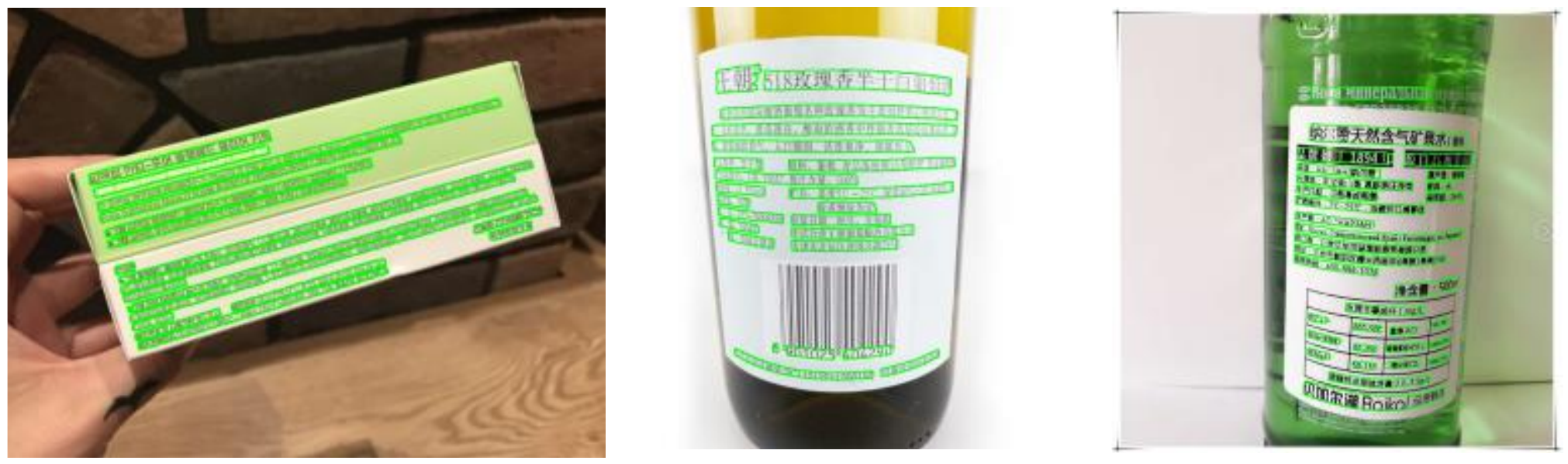} 
\caption{Detection results on DAST1500.}
\label{fig9}
\end{figure*}

\textbf{Dense adjacent text detection.} As shown in Table \ref{t10}, we compare the proposed method with state-of-the-art methods on DAST1500. For the F-measure, the method surpasses the previous state-of-the-art method by 1.9\%. The performance on DAST1500 demonstrates the solid superiority of the proposed method to detect dense and arbitrary-shaped scene text.

From all the Table results, there is a clear trend that the performance of the proposed method is improved by using extra data during training, compared to the results without using extra data. There may be some reasons: on the one hand, according to the statistical learning theory, in order to get a model of small test error, we need to have more data to suppress the model complexity penalty. On the other hand, more data can contain various scene texts, which can improve the robustness of the deep learning model.

\subsection{Speed}
Although our method is segmentation-based, we take the speed of the entire pipeline into consideration. Optimized along with the proposed text region representation and the polygon expansion algorithm, the segmentation network can not only simplify the post-processing but also enhances the performance of arbitrary-shaped text detection. The speed of the proposed method is compared with two other methods as shown in Table \ref{t11}, which are all able to deal with arbitrary shape scene text. From the table, we can see that the speed of our method is much faster than the other two methods. While complicated post-processing is needed in PSENet\cite{wang2019shape} and TextSnake\cite{long2018textsnake}, the post-processing in the proposed method is efficient.

\subsection{Qualitative results}
Figure \ref{fig7} illustrates qualitative results on CTW1500 and TotalText. Figure \ref{fig8} shows some examples of MSRA-TD500 and on ICDAR2015. And Figure \ref{fig9} is for DAST1500. We can find that our proposed method is adaptive to texts with arbitrary shapes, including horizontal text, multi-oriented text and curved text.

\section{Conclusion}
In this paper, we propose a robust deformable framework with novel text region representation to successfully detect the text instances of arbitrary shapes in natural images. And with the training strategy, we can improve the representation robustness against text instances of various scales and aspect ratios. By using expanding ratios to expand the central text regions with our proposed polygon expansion algorithm, we can detect the text instances effectively and it is quite easy to separate spatially close text instances. The performance on scene text detection benchmarks demonstrates the effectiveness of the text region representation and the post-processing algorithm. Possible future work includes extending our method to other segmentation tasks, addressing challenges of arbitrary oriented text detection and shortening the segmentation-based methods’ running time.

\begin{table}[H]
\centering
\resizebox{0.6\columnwidth}{!}{
\smallskip\begin{tabular}{|l|l|l|}
\hline
Method &  Scale & Speed \\
\hline
TextSnake\cite{long2018textsnake} & 768 & 1.1fps \\
\hline
PSENet\cite{wang2019shape} & 736 & 3.8fps \\
\hline
Ours & 736 & 8.6fps \\
\hline
\end{tabular}
}
\caption{Speed compared on different detection methods supporting arbitrary shape texts on IC15.}
\label{t11}
\end{table}

\input{text_detection.bbl}

\bibliographystyle{IEEEtran}
\bibliography{text_detection}

\EOD
\end{document}